\documentclass[letterpaper]{article} 
\usepackage{aaai25}  
\usepackage{times}  
\usepackage{helvet}  
\usepackage{courier}  
\usepackage[hyphens]{url}  
\usepackage{graphicx} 
\usepackage{natbib}  
\usepackage{caption} 
\usepackage{algorithm}
\usepackage{algorithmic}
\usepackage{amssymb}
\usepackage{amsmath}
\usepackage{graphicx}
\usepackage{subfigure}
\usepackage{enumitem}
\usepackage{pifont}
\usepackage{booktabs}
\usepackage{bm}
\newcommand{\xmark}{\ding{55}}

\usepackage{multirow}

\newtheorem{prop}{Proposition}

\newlist{Properties}{enumerate}{2}
\setlist[Properties]{label=Property \arabic*., font=\textbf, itemindent=*}

\DeclareMathOperator*{\argmin}{argmin}
\usepackage{newfloat}
\usepackage{listings}
\DeclareCaptionStyle{ruled}{labelfont=normalfont,labelsep=colon,strut=off} 
\floatstyle{ruled}
\newfloat{listing}{tb}{lst}{}
\floatname{listing}{Listing}
\nocopyright

\title{Efficient Image-to-Image Diffusion Classifier for Adversarial Robustness}

\author {
    Hefei Mei\textsuperscript{\rm 1},
    Minjing Dong\thanks{Corresponding Author}\textsuperscript{\rm 1},
    Chang Xu\textsuperscript{\rm 2}
}
\affiliations {
    \textsuperscript{\rm 1}City University of Hong Kong, China\\
    \textsuperscript{\rm 2}University of Sydney, Australia\\
    hefeimei2-c@my.cityu.edu.hk, minjdong@cityu.edu.hk, c.xu@sydney.edu.au
}

\usepackage{bibentry}

\begin{document}

\maketitle

\begin{abstract}

Diffusion models (DMs) have demonstrated great potential in the field of adversarial robustness, where DM-based defense methods can achieve superior defense capability without adversarial training. However, they all require huge computational costs due to the usage of large-scale pre-trained DMs, making it difficult to conduct full evaluation under strong attacks and compare with traditional CNN-based methods. 
Simply reducing the network size and timesteps in DMs could significantly harm the image generation quality, which invalidates previous frameworks. To alleviate this issue, we redesign the diffusion framework from generating high-quality images to predicting distinguishable image labels. Specifically, we employ an image translation framework to learn many-to-one mapping from input samples to designed orthogonal image labels. Based on this framework, we introduce an efficient Image-to-Image diffusion classifier with a pruned U-Net structure and reduced diffusion timesteps. 
Besides the framework, we redesign the optimization objective of DMs to fit the target of image classification, where a new classification loss is incorporated in the DM-based image translation framework to distinguish the generated label from those of other classes. We conduct sufficient evaluations of the proposed classifier under various attacks on popular benchmarks. Extensive experiments show that our method achieves better adversarial robustness with fewer computational costs than DM-based and CNN-based methods. The code is available at https://github.com/hfmei/IDC
\end{abstract}

%

\section{Introduction}

Diffusion models (DMs) have achieved excellent performance in high-quality generative tasks~\cite{ho2020denoising, song2020score, song2020denoising, dhariwal2021diffusion, rombach2022high, ma2024followyourpose, ma2024followyourclick, ma2024followyouremoji}, leading to a range of relative applications, such as in-painting~\cite{lugmayr2022repaint, zhang2023sine}, audio or video synthesis~\cite{ho2022imagen, singer2022make, wang2024cove, chen2024follow}, image super-resolution~\cite{saharia2022image, li2022srdiff} and image-to-image translation~\cite{saharia2022palette, sasaki2021unit, li2023bbdm}.

Adversarial robustness refers to the capability of a model to maintain its accuracy when faced with deliberately crafted adversarial inputs intended to deceive it~\cite{rice2020overfitting, liu2018towards, dong2023adversarial, kang2024diffattack, dong2022random}. Previous defense methods mainly focus on adversarial training \cite{madry2017towards} which involves generated adversarial examples during optimization. On the contrary, DM-based adversarial defense methods mainly utilize the inherent defense ability of DMs, such as adversarial purification that applies DMs to remove perturbations~\cite{nie2022diffusion} and robust diffusion classifier (RDC) that treats DMs as naturally robust generative classifiers~\cite{chen2023robust}. These DM-based methods have demonstrated superior performance in achieving adversarial robustness compared with CNNs.

\begin{figure}
\centerline{\includegraphics[width=0.98\linewidth]{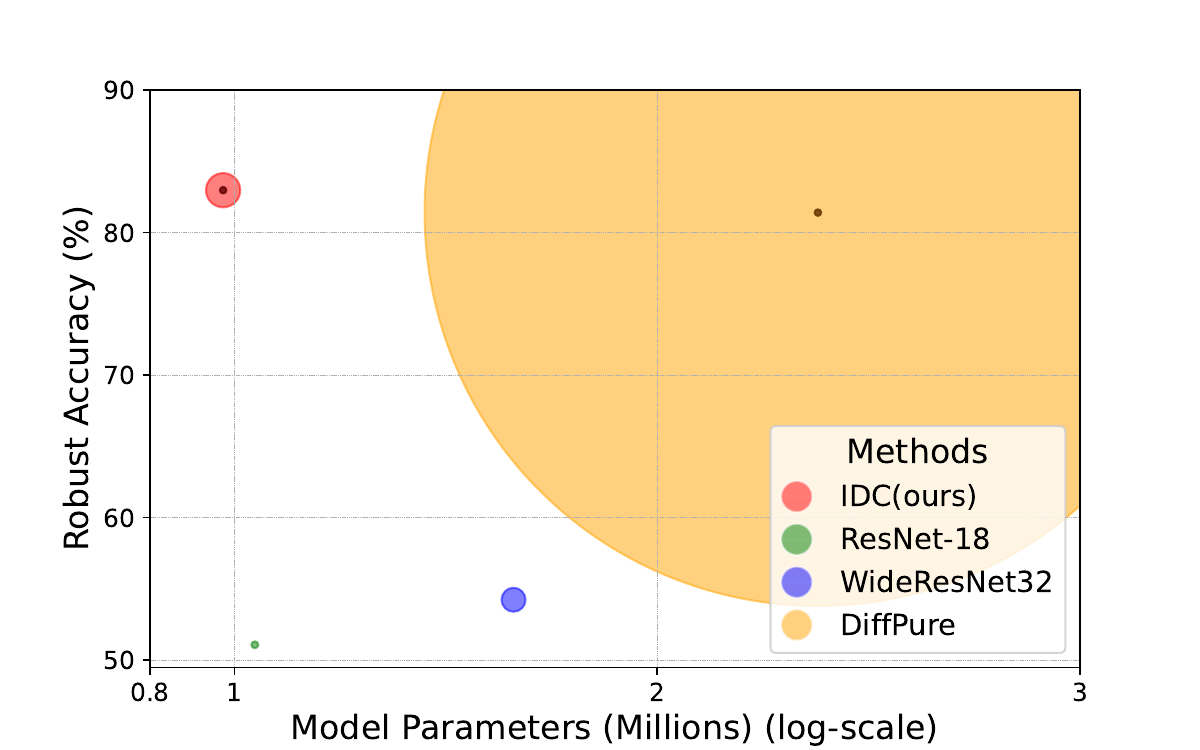}}
    \caption{Comparison of robust accuracy with both CNN-based and DM-based benchmarks against BPDA+EOT attack. The area of a circle demonstrates the inference FLOPs where the FLOPs of our introduced IDC are 27.09G and those of DiffPure are 14.26T. The dark center of the circle is located at the accurate value.}
    \label{f-introparam}
\end{figure}

Despite their advanced defense capability, the computational expense of these methods could be rather high, which makes it expensive to deploy them in real-world applications or even conduct a full evaluation on the entire test dataset compared with traditional CNN-based methods.
For example, RDC~\cite{chen2023robust} adopts a Zero-shot Diffusion Classifier (ZDC)~\cite{li2023your} as the prediction paradigm, which relies on a large-scale pre-trained DM and needs to iteratively perform a diffusion process for all classes to derive class distribution. 
DiffPure~\cite{nie2022diffusion} connects pre-trained DM with traditional classifier~\cite{he2016deep, zagoruyko2016wide}, where DM serves as a purification transformation of potential adversarial examples. However, its computation during the inference stage is boosted due to the complexity of large U-Net~\cite{ronneberger2015u}, diffusion timesteps, and an additional classifier. As illustrated in Figure \ref{f-introparam}, the FLOPs of ResNet-18~\cite{he2016deep} are $1.12$G, while those of DiffPure are $14.26$T.
Thus, we are motivated to explore a more efficient manner to incorporate diffusion models in adversarial robustness. A naive method is to reduce the size of the U-Net and decrease the number of diffusion timesteps, however, such simplification conflicts with the objective of high-quality generation in DMs which could play an important role in DM-based methods.
We mainly attribute this issue to the gap between the generation task in DMs and the classification task in adversarial robustness.

In this paper, we aim to redesign the diffusion framework for adversarial robustness to eliminate this gap, relieving the pressure on DMs to generate high-quality images, which enables an efficient incorporation of DMs in adversarial robustness. Specifically, we regard classification tasks as image translation tasks achieved by DMs, where the image is translated to the pre-defined labels that are presented by different images. Since the objective is to align images with their corresponding image labels instead of high-quality image generation, the required complexity of DMs in adversarial robustness can be significantly reduced. With this simple framework, we introduce an Image-to-Image Diffusion Classifier (IDC) to effectively and efficiently apply DMs to adversarially robust image classification.
In order to guarantee the diversity of pre-defined image labels for classification, we propose to construct orthogonal image labels in pixel space. Then IDC employs an image-to-image translation framework to learn a many-to-one mapping from input images to these orthogonal image labels, and the classification can be achieved by measuring the distances between translated results and all pre-defined image labels. Due to the switch of objectives, we are able to alleviate the computation of DMs in different aspects, including pruning the U-Net structure and reducing the diffusion timesteps. Furthermore, the vanilla optimization of translation frameworks only minimizes the distance between the translated result and its corresponding image labels without considering other classes. Thus, we incorporate a classification loss in the optimization of DM-based image translation to better fit the classification objective. Our introduced IDC achieves superior performance in the trade-offs between model efficiency and adversarial robustness. As shown in Figure \ref{f-introparam}, we achieve more competitive robustness with fewer model parameters than CNN-based adversarial training methods and much lower FLOPs cost than the DM-based method.

We summarize the main contributions as follows:
\begin{itemize}[leftmargin=*]
\item  We propose a novel Image-to-Image diffusion classifier with pre-defined orthogonal image labels, which naturally converts image generation tasks to classification tasks. 
\item Through task conversion, we propose to reduce the complexity of diffusion models without harming the performance, including network size and diffusion steps.
\item We propose a classification loss during the optimization of IDC, which improves the adaptability of the image translation framework for classification tasks.
\item We perform extensive experiments to empirically demonstrate the superiority of IDC on various benchmarks.
\end{itemize}

\section{Related Work}
\label{relatedwork}

\subsection{Diffusion Models}

As one of the advanced probabilistic generative models, Diffusion Models (DMs)~\cite{ho2020denoising, song2020denoising, song2020improved, dhariwal2021diffusion} employ a parameterized  Markovian chain to estimate the target distribution, which has been used in many relative tasks, such as diffusion classification (DC)~\cite{li2023your, chen2023robust}, object detection~\cite{chen2023diffusiondet} and image-to-image translation~\cite{saharia2022palette, li2023bbdm}. In this paper, we are concerned about the high potential of DMs in adversarial robustness.

\subsection{CNN-based Adversarial Defense}

The core strategies for CNN-based defense methods mainly focus on adversarial training (AT)~\cite{madry2017towards, zhang2019theoretically, dong2022random, lin2024adversarial} and adversarial purification (AP)~\cite{samangouei2018defense, grathwohl2019your, hill2020stochastic}. The former enhances model robustness by integrating adversarial examples into the training process while the latter focuses on removing adversarial noise from input samples. 

\subsection{DM-based Adversarial Defense}

The existing diffusion classifier~\cite{li2023your} calculates probabilities by applying Bayes' theorem to the outputs of a conditional diffusion model while RDC~\cite{chen2023robust} enhances its adversarial robustness. Diffusion purification methods~\cite{nie2022diffusion, zhang2023purify++, wang2022guided, wu2022guided, lin2024robust} purify adversarial samples through DMs before passing them to a cascaded general classifier. Distinct from prior research, we design an efficient DM-based robust classifier for the entire dataset evaluation. 

\section{Methodology}
\label{methodology}

In this section, we explore the application of diffusion models in adversarial robustness. Specifically, we start with the preliminary of the diffusion model. Then we discuss the limitations of existing diffusion classifiers as well as diffusion models for adversarial robustness, which motivates us to introduce an efficient image-to-image diffusion classifier (IDC) to tackle the aforementioned problems. To further eliminate the gap between the diffusion-based optimization objective in IDC and the classification objective, a classification loss is thus incorporated into the loss function.

\subsection{Preliminary}
\label{preliminary}

\subsubsection{Diffusion Model}

Diffusion models~\cite{ho2020denoising, song2020denoising} consider the generative task as a Markovian chain process. In the forward process, the DM adds Gaussian noise to real data $\boldsymbol{y}_0$ by $T$ timesteps, which converts $\boldsymbol{y}_0$ to Gaussian noise $\boldsymbol{y}_T \sim \mathcal{N}(0, 1)$. The diffusion forward process at each step can be defined as $\boldsymbol{y}_t = \sqrt{1-\beta_t}\boldsymbol{y}_0+\sqrt{\beta_t}\boldsymbol{\epsilon}_t$, where $\beta_t$ is a linearly increased scale factor, $\boldsymbol{\epsilon}_t \sim \mathcal{N}(0, 1)$. To simplify the calculation, the forward process can be formulated as:
\begin{equation}
    \boldsymbol{y}_t = \sqrt{\overline{\alpha}_t}\boldsymbol{y}_0+\sqrt{1-\overline{\alpha}_t}\boldsymbol{\epsilon},
    \label{y0_to_yt_ddpm}
\end{equation}
where $\alpha_t = 1 - \beta_t, \overline{\alpha}_t=\prod \limits\nolimits_{s=1}^t \alpha_s$. In the reverse process, the DM is trained to denoise the sampled Gaussian noise $\boldsymbol{y}_{t}$ to $\boldsymbol{y}_{t-1}$, with which the generation of data $\hat{\boldsymbol{y}}_{0}$ follows another Markovian chain process. The reverse process of DM in timestep $t$ can be formulated as:
\begin{equation}
\begin{aligned}
 & p_\theta(\boldsymbol{y}_{t-1}\mid \boldsymbol{y}_t)=\mathcal{N}(\boldsymbol{y}_{t-1} ; \mu_{\theta}(\boldsymbol{y}_t, t), \delta_t \boldsymbol{I}),\\
 & \mu(\boldsymbol{y}_t, t) = \frac{1}{\sqrt{\alpha_t}}(\boldsymbol{y}_t- \frac{\beta_t}{\sqrt{1-\bar{\alpha}_t}} \epsilon_{\theta}(\boldsymbol{y}_t, t)),\\
\end{aligned}
\end{equation}
where $\epsilon_{\theta}$ is a noise predictor parameterized by $\theta$. It is a common practice for DMs to utilize a U-Net~\cite{ronneberger2015u} as the noise predictor, where $\theta$ can be optimized by maximizing the lower variation limit. The training loss of DMs can be formulated as:
\begin{equation}
\min _{\theta} \mathbb{E}_{t \sim U[0,1], \boldsymbol{y}_0, \boldsymbol{\epsilon} \sim \mathcal{N}(\mathbf{0}, \mathbf{I})} \left\|\epsilon_{\theta}\left(\boldsymbol{y}_t, t\right)-\boldsymbol{\epsilon} \right\|_2^2.
\end{equation}

\subsubsection{Diffusion Classifier}

Zero-shot Diffusion Classifier~\cite{li2023your} performs training-free classification using the inherent capabilities of Diffusion model $U_L$. As shown in Figure \ref{f-zdc}, the core of ZDC is the application of Bayes' theorem to the outputs of a conditional generative model, combining the prediction likelihoods with prior knowledge of class distributions to calculate the posterior probability. For the input image $\boldsymbol{x}$ and a class label $c_i$, the ZDC utilizes the Evidence Lower Bound (ELBO) in place of $\text{log} p_\theta(\boldsymbol{x} \mid c)$ then reformulated the posterior distribution as:
\begin{equation}
p_\theta(c_i \mid \boldsymbol{x})=\frac{\exp \{-\mathbb{E}_{t, \epsilon}[\|\epsilon_\theta(\boldsymbol{x}_t, c_i)-\boldsymbol{\epsilon}\|^2]\}}{\sum_j \exp \{-\mathbb{E}_{t, \epsilon}[\|\epsilon_\theta(\boldsymbol{x}_t, c_j)-\boldsymbol{\epsilon}\|^2]\}},
\end{equation}
where the iteration $j \in [1, 2, 3, ..., K]$, and $K$ is the number of the classes.

\subsubsection{Diffusion Purification}

As shown in Figure \ref{f-diffpure}, by combining the diffusion model with the classifier, Diffusion Purification (DiffPure)~\cite{nie2022diffusion} achieves considerable adversarial robustness. During purification, the adversarial sample $\boldsymbol{x}'$ is diffused to the timestep $t_L$, which can be sampled as $\boldsymbol{x}(t_L)=\sqrt{\alpha(t_L)}\boldsymbol{x}'+\sqrt{1-\alpha(t_L)}\boldsymbol{\epsilon}$. Then the sample $\boldsymbol{x}'$ is purified through the reverse process by U-Net $U_L$ and subsequently input into a CNN-based classifier to obtain the results, which can be formulated as: 
\begin{equation}
p_{\theta_U,\theta_f}(c \mid \boldsymbol{x})= f\left(\underset{t={1...t_L}}{\text{reverse}} \ U_L(\boldsymbol{x}(t_L), t)\right),
\end{equation}
where the adversarial sample $\boldsymbol{x'}$ need to be purified by the $t_L$-timestep diffusion model and classified by classifier $f$.

\subsection{Image-to-Image Diffusion Classifier}
\label{classifier}

\begin{figure}
    \centering
    \subfigure[Zero-shot diffusion classifier (ZDC)]{
    \includegraphics[width=1.0\linewidth]{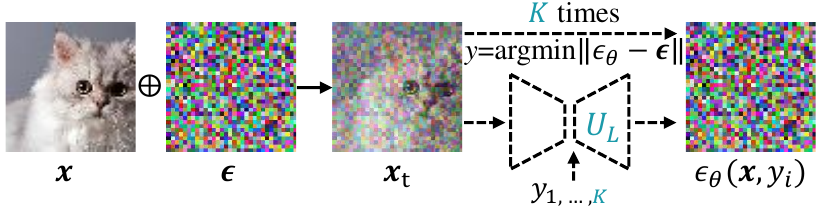}
        \label{f-zdc}       
    }
    \subfigure[Diffusion for purification (DiffPure)]{
	\includegraphics[width=1.0\linewidth]{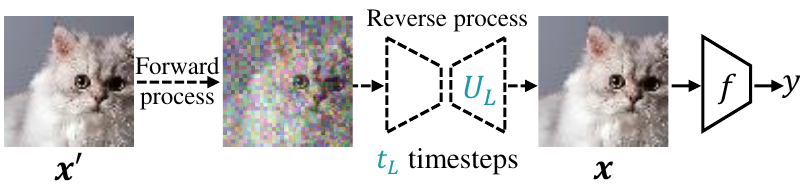}
        \label{f-diffpure}   
    }
    \subfigure[Image-to-Image diffusion classifier (IDC)]{
	\includegraphics[width=0.86\linewidth]{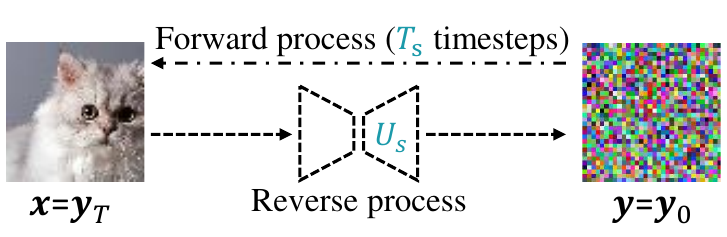}
        \label{f-idc}
    }
    \caption{Comparison of different diffusion classifier paradigms. The number of iterations $K>T_s$ in our IDC while the timesteps $t_L \gg T_s$. The parameters of $U_L$ in both ZDC and DiffPure are also larger than $U_s$ in IDC.}
    \label{f-labelset}
\end{figure}

\begin{figure*}
\centerline{\includegraphics[width=1.0\linewidth]{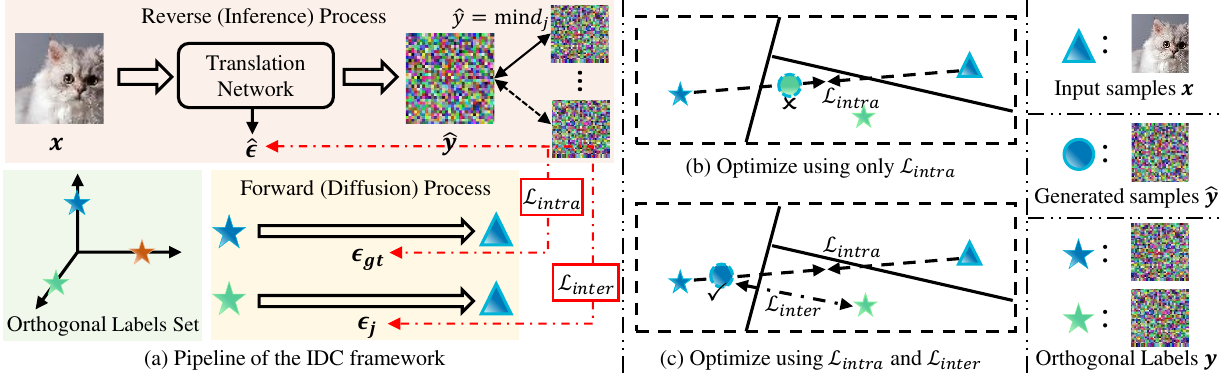}}
	\caption{The illustration of our framework and optimization loss. Triangles represent input samples, while circles represent generated samples of the network. In Figure (a), we represent orthogonal labels in a high-dimensional pixel space using a three-dimensional schematic. The differently colored pentagrams each correspond to image labels of distinct categories.}
    \label{f-framework}
\end{figure*}

Applying a pre-trained diffusion model to classification tasks requires additional components or computations. For example, diffusion classifiers~\cite{chen2023robust,li2023your} perform inference for each class separately, which leads to a computational overhead that scales proportionally with the number of classes $K$, as shown in Figure \ref{f-zdc}. DiffPure~\cite{nie2022diffusion} requires an additional pre-trained classifier and $t_L$ steps of the diffusion process, as shown in Figure \ref{f-diffpure}. All the iterative steps in these methods heavily rely on the large-scale U-Net~\cite{ronneberger2015u}, which makes their inference phases much slower than traditional CNN-based classifiers. 

Given these limitations, we are motivated to alleviate huge computational costs in DMs for adversarial robustness, such as the size of U-Net and the number of diffusion steps.
However, reducing computation in diffusion models could significantly influence the image generation quality, which invalidates previous frameworks, such as purification in~\cite{nie2022diffusion} and noise predictor in \cite{chen2023robust}. Thus, we propose to redesign the diffusion framework for adversarial robustness, which relieves the demand for high-quality image generation.

\subsubsection{Orthogonal Image Labels Generation}
\label{labelset}

Different from the pre-trained DM classifier of ZDC~\cite{li2023your}, we employ the image translation task of DMs to the classification task, which converts the high-quality image generation into image label alignment. To achieve this, we first need to construct image labels for classification and then apply the translation framework to the classification task.
Considering the labels $y \in \mathbb{N}^b$ for traditional classifiers ~\cite{simonyan2014very, he2016deep, dosovitskiy2020image}, which maintain the same data distance and orthogonality, we construct orthogonal image labels $\boldsymbol{y} \in \mathbb{R}^{[b, c, h, w]}$ in the pixel space.
To this end, we first generate a random noise vector $V \in \mathbb{R}^{[K, c*h*w]}$ in the space formed by classes and pixels where $K$ is the number of classes. The orthogonal vectors can be obtained by performing QR decomposition~\cite{francis1961qr} on random vectors $V$, represented as $V=QR$, where $Q \in \mathbb{R}^{[K, c*h*w]}$ is an orthogonal matrix. By dividing and resizing the matrix $Q$, we can obtain the image label $\boldsymbol{y}^i \in \mathbb{R}^{[c, h, w]}$ for each class. Next, we normalize the images to ensure that the pixel ranges of the image labels and the inputs are consistent. During training, $\boldsymbol{y}^i$ can form a batch of image labels $\boldsymbol{y} \in \mathbb{R}^{[b, c, h, w]}$ based on original labels $y \in \mathbb{N}^{b}$. 

\subsubsection{Image Label Translation}

After constructing the image labels, we adopt the image translation framework BBDM~\cite{li2023bbdm} to realize the image translation from the inputs to labels as shown in Figure \ref{f-idc}. Compared with ZDC~\cite{li2023your} in Figure \ref{f-zdc}, which has a computational complexity of $O(K)$ for the number of classes, our framework achieves $O(1)$ complexity for the classes during reference. In the forward process, the image labels $\boldsymbol{y}=\boldsymbol{y}_0$ progressively introduce Gaussian noise, generating a series of increasing noise samples $\boldsymbol{y}_1, \boldsymbol{y}_2, \ldots, \boldsymbol{y}_T$, which can be expressed as (specified in Appendix A.1):
\begin{equation}
    \boldsymbol{y}_t = (1-m_t)\boldsymbol{y}_0+m_t\boldsymbol{x}+\sqrt{\delta_t}\boldsymbol{\epsilon}_t
    \label{y0_to_yt},
\end{equation}
\begin{equation}
\begin{aligned}
q(\boldsymbol{y}_t\mid \boldsymbol{y}_{t-1}, \boldsymbol{x}) = \mathcal{N}&(\boldsymbol{y}_t; \gamma \boldsymbol{y}_{t-1}+(m_t-\gamma m_{t-1})\boldsymbol{x}, \\
&(\delta_{t} - \gamma^2 \delta_{t-1})\boldsymbol{I}),
\label{xt-1_to_xt}
\end{aligned}
\end{equation}
where $\gamma=\frac{1-m_t}{1-m_{t-1}}$, Eq. (\ref{y0_to_yt}) represents the forward process from $\boldsymbol{y}_0$ to $\boldsymbol{y}_t$, $\delta_t$ denotes the variance of diffusion process, $\boldsymbol{\epsilon}_t \sim \mathcal{N}(\boldsymbol{0}, \boldsymbol{I})$.
In the reverse process, the DM learns the denoising process from input image $\boldsymbol{x}=\boldsymbol{y}_T$ to image label $\boldsymbol{y} = \boldsymbol{y}_0$. For timestep $t$, the reverse process of transition can be expressed as $q(\boldsymbol{y}_{t-1}\mid \boldsymbol{y}_t, \boldsymbol{y}_0, \boldsymbol{x})=\mathcal{N}(\boldsymbol{y}_{t-1};\mu(\boldsymbol{y}_{t}, \boldsymbol{x}), \delta \boldsymbol{I})$ where the mean value $\mu$ is the noise predictor, and the prediction of the DM can be expressed as $p_\theta(\hat{\boldsymbol{y}}_{t-1}\mid \boldsymbol{y}_t, \boldsymbol{x})=\mathcal{N}(\hat{\boldsymbol{y}}_{t-1}; \hat{\mu}_\theta(\boldsymbol{y}_t, \boldsymbol{x}, t), \delta_t \boldsymbol{I})$ where $\theta$ is the parameter of the network optimized by $\mu_\theta$. 

\subsubsection{Image-to-Image Classification}

Given the timestep $T_s$ and the input image $\boldsymbol{x}=\boldsymbol{y}_T$, the distribution $p_\theta(\hat{\boldsymbol{y}}_0)$ can be generated following the Markovian chain of reverse process $p_\theta(\hat{\boldsymbol{y}}_{t-1}\mid \boldsymbol{y}_t, \boldsymbol{x})$.
After sampling the translated image label $\hat{\boldsymbol{y}}_0$, the predicted class $c$ can be derived by the distance between $\hat{\boldsymbol{y}}_0$ and all pre-defined image labels, which is formulated as (specified in Appendix A.3):
\begin{equation}
c = \mathop{\arg\min}\limits_{i} \Vert \hat{\boldsymbol{y}}_0 - \boldsymbol{y}^i_0 \Vert_1,
\label{pre_y}
\end{equation}
where $\boldsymbol{y}^i_0$ denotes the pre-defined image label of class $i$.

\subsection{Diffusion Complexity Reduction}
\label{streamlining}

Although we reduce the computational complexity from $O(K)$ to $O(1)$ compared with ZDC~\cite{li2023your}, the large U-Net\cite{ronneberger2015u} structure and long timesteps still cause a computational burden during robustness evaluation. 
Fortunately, our framework transforms the task from generating high-quality images to aligning image labels. This simplification of the objective allows for a certain decrease in image generation quality, providing an opportunity to simplify the diffusion process by pruning the U-Net structure and reducing the diffusion timesteps.
\subsubsection{Network Structure Pruning} 
At the beginning of U-Net architecture~\cite{ronneberger2015u}, there is a convolutional layer to convert the channel of the input image to the model channel $C_m$. As the model channel directly affects all channels of the network, we reduce the original U-Net model channel to  $c_m=C_m/k$, where $k$ is used to adjust the degree of network pruning. 
Given the contracting path of U-Net architecture, it consists of several unpadded convolutions for downsampling and upscales the feature channels by a factor of $u_1$ while the overall upscale list can be $u=[u_1, u_2, ..., u_n]$. By adjusting it to lower multiplication factors, parameters in the corresponding unpadded convolutions can be effectively reduced.
Finally, we cut the number of ResNet blocks $n_R$ between two unpadded convolutions, which achieves a further decrease in parameters. The structure of the expansive path in the U-Net is adapted to match the adjusted contracting path. Through the pruning methods, we reduce the parameters from the initial BBDM model of 237.09M to 9.39M (specified in Appendix C.1).
\subsubsection{Diffusion Timestep Reduction} 
To achieve high-quality image generation, hundreds or even thousands of timesteps are always required for diffusion models~\cite{ho2020denoising, song2020denoising, dhariwal2021diffusion}, which results in a significant computational cost for the adversarial robustness evaluation.
In our framework, the image generation process is greatly simplified, allowing the network to learn the diffusion process with only a few timesteps. As shown in Figure \ref{f-labelset}, we reduce the general diffusion timesteps $t_L$ to $T_s$ where $t_L \gg T_s$, greatly reducing the FLOPs of the inference process.

\subsection{Classification Optimization of Diffusion Classifier}
\label{s-contloss}
Thanks to the switch from high-quality image generation to image label translation, the complexity of the diffusion process can be significantly alleviated. However, unlike traditional CNN-based classifiers, the proposed image-to-image diffusion classifier cannot be optimized by the popular classification loss, such as cross-entropy, due to the diffusion-based framework. Thus, we need to reformulate the classification optimization of diffusion classifiers. We mainly divide the classification optimization into two parts, including the intra-class loss which minimizes the distance between the translated image label $\hat{\boldsymbol{y}}^i_0$ and the ground truth $\boldsymbol{y}^i_0$, and the inter-class loss which maximizes the distance between the translated image label $\hat{\boldsymbol{y}}^i_0$ and those of other classes $\boldsymbol{y}^j_0$ where $j \neq i$. 
First, we discuss the intra-class loss in the context of the diffusion classifier. It is obvious that the optimization of intra-class loss is naturally achieved by the training objective ELBO in the image-to-image translation framework, which is formulated as
\begin{equation}
\displaystyle
\scalebox{0.92}{$
\begin{aligned}
- &\mathbb{E}_q \Big[\sum_{t=2}^{T_s} D_{KL}(q(\boldsymbol{y}^i_{t-1}\mid \boldsymbol{y}^i_t, \boldsymbol{y}^i_0, \boldsymbol{x})\Vert p_\theta(\hat{\boldsymbol{y}}^i_{t-1}\mid \boldsymbol{y}^i_t, \boldsymbol{x})) \\
&- \log p_\theta(\hat{\boldsymbol{y}}^i_0\mid \boldsymbol{y}^i_1, \boldsymbol{x}) + D_{KL}(q(\boldsymbol{y}^i_{T}\mid \boldsymbol{y}^i_0, \boldsymbol{x})\Vert p(\boldsymbol{y}^i_{T}\mid \boldsymbol{x}) \Big],
\label{ELBO_intra}
\end{aligned}
$}
\end{equation}
where the last term can be ignored as $\boldsymbol{y}^i_{T}$ is equal to $\boldsymbol{x}$. Based on the reparametrization method~\cite{ho2020denoising, li2023bbdm}, the above training objective can be simplified as the following loss (specified in Appendix A.2):
\begin{equation}
\mathcal{L}_{intra} = \Vert m_t(\boldsymbol{x}-\boldsymbol{y}^i_0) + \sqrt{\delta_t}\boldsymbol{\epsilon}_t -\epsilon_\theta(\boldsymbol{y}^i_t, t) \Vert_1.
\label{loss_intra}
\end{equation}
Besides the intra-class in Eq. (\ref{loss_intra}), the inter-class loss of the diffusion classifier remains unexplored. Thus, we propose to design a loss $\mathcal{L}_{inter}$ to push the translated image label $\hat{\boldsymbol{y}}^i_0$ away from other classes, as shown in Figure \ref{f-framework}(c).
Formally, we consider the most confusing class $j$ of image $\boldsymbol{x}$ where the image label $\boldsymbol{y}^j_0$ is the most closest one to the translated $\hat{\boldsymbol{y}}^i_0$, as $j=\argmin_j \Vert \hat{\boldsymbol{y}}^i_0 - \boldsymbol{y}^j_0 \Vert_1$ and $j \neq i$. In the framework of translation, the distance between $\hat{\boldsymbol{y}}^i_0$ and the most confusing image label $\boldsymbol{y}^j_0$ is expected to be maximized, which can be formulated in a simplified reverse format of ELBO as:
\begin{equation}
\displaystyle
\scalebox{0.86}{$ - \mathbb{E}_q \Big[-\sum_{t=2}^{T_s} D_{KL}(q(\boldsymbol{y}^j_{t-1}\mid \boldsymbol{y}^j_t, \boldsymbol{y}^j_0, \boldsymbol{x})\Vert p_\theta(\hat{\boldsymbol{y}}^i_{t-1}\mid \boldsymbol{y}^i_t, \boldsymbol{x})) \Big],
$}
\label{ELBO_inter}
\end{equation}
where the constant is ignored. Similar to the derivation in DDPM \cite{ho2020denoising}, the term $q(\boldsymbol{y}^j_{t-1} | \boldsymbol{y}^j_t, \boldsymbol{y}^j_0, \boldsymbol{x})$ can be derived through Bayes' theorem and formulated as $\mathcal{N}(\boldsymbol{y}^j_{t-1};\mu(\boldsymbol{y}^j_{t}, \boldsymbol{y}^j_{0}, \boldsymbol{x}), \delta \boldsymbol{I})$. Through training a network to estimate the noise $\epsilon_\theta$ in the mean value term $\mu$, a simplified version of the ELBO objective in Eq. (\ref{ELBO_inter}) can be derived, as shown in the following proposition.

\begin{prop}
\textit{
For the objective in the Eq. (\ref{ELBO_inter}), the training loss could be simplified as:
}
\begin{equation}
\scalebox{1.0}{$
\begin{aligned}
    \mathcal{L}_{inter} = -\Vert & m_t(\boldsymbol{x}-\boldsymbol{y}^j_0) + m_{t-1}(\boldsymbol{y}^j_0-\boldsymbol{y}^i_0) \\
    & + \sqrt{\delta_t}\boldsymbol{\epsilon}_t -\epsilon_\theta(\boldsymbol{y}^i_t, t) \Vert_1.\\
\end{aligned}
$}
\label{loss_inter}
\end{equation}
\label{prop}
\end{prop}
The detailed proof of Proposition \ref{prop} is provided in Appendix B.1. Finally, the overall objective loss of the IDC can be formulated as $\mathcal{L}_{cls} = \mathcal{L}_{intra} + \alpha \mathcal{L}_{inter}$, where $\alpha$ is a hyper-parameter to balance the influence of $\mathcal{L}_{inter}$.

\section{Experiments}

\subsection{Experimental Settings}

\subsubsection{Datasets and Metrics.}

We conduct experiments on the CIFAR-10, CIFAR-100 datasets~\cite{krizhevsky2009learning} and Tiny-ImageNet~\cite{deng2009imagenet}. We compare our classifier with CNN-based and DM-based methods, using two metrics: standard accuracy and robust accuracy. Different from the setting of most DM-based methods, which randomly select a test subset for robust evaluation, we evaluate the performance of our IDC on the entire test dataset. 

\subsubsection{Adversarial Attacks.}

We evaluate our methods with CNN-based methods under five attacks: PGD~\cite{madry2017towards}, FGSM~\cite{szegedy2013intriguing}, MIFGSM~\cite{dong2017discovering}, CW~\cite{carlini2017towards} and AutoAttack~\cite{croce2020reliable}. For each attack, $\epsilon$ is set to 8/255, the attack steps of PGD and MIFGSM are set to 20 and 5, and the steps and learning rate of CW are 1000 and 0.01. Following \cite{lee2023robust, nie2022diffusion}, we evaluate DM-based method under BPDA+EOT attack~\cite{hill2020stochastic} with $\ell_{\infty}$ perturbations and PGD+EOT attack~\cite{athalye2018obfuscated} while $\epsilon=8/255$, steps are 200 and EOT is set to 20.

\subsubsection{Diffusion Classifier Setup.}

We train our classifier using the Adam optimizer with 256 batch size cross 4 Tesla V100-32GB GPUs, CUDA V10.2 in PyTorch V1.7.1~\cite{paszke2019pytorch}. The diffusion timesteps are set to $T_s=4$, and the learning rate is set to 0.0001. For the CIFAR-10 dataset, we train 400 epochs in total while we train 600 epochs for the CIFAR-100 dataset. The hyper-parameter $\alpha$ is set to 0.2. The model channels of the U-Net are set to $c_m=64$, and the number of ResNet blocks is set to $n_R=1$. For the CIFAR-10 dataset, the upscale list of channels is set to $u=[1,4]$ while that for CIFAR-100 is set to $u=[1,4,8]$.

\begin{table*}
    \centering
  \begin{tabular}{c|c|c|c|c|c|c|c|c|c}
    \toprule
    Dataset & Method &  Params & AT & Standard & $\text{PGD}^{20}$ & FGSM & MIFGSM & CW & AutoAttack \\
    \midrule
    \multirow{5}*{CIFAR-10} & \multirow{2}*{ResNet-18} & \multirow{2}*{11.17M} & \xmark & 95.0 & 0.0  & 43.8 & 2.1 & 66.8 & 0.0 \\
     &  &  & \checkmark & 82.7 & 51.5 & 57.3 & 55.0 & 78.9 & 48.5 \\ 
     \cmidrule{2-10}
      & \multirow{2}*{WideResNet32} & \multirow{2}*{46.16M} & \xmark & 96.1 & 0.0 & 51.5 & 3.9 & 69.7 & 0.0 \\
     &  &  & \checkmark & 86.6 & 54.7 & 61.4 & 58.8 & \textbf{83.4} & 52.5 \\
     \cmidrule{2-10}
     &  IDC(Ours) & 9.39M & \xmark & 85.9 & \textbf{84.8} & \textbf{83.6} & \textbf{84.4} & 81.9 & \textbf{59.9} \\
    \midrule
    \multirow{5}*{CIFAR-100} & \multirow{2}*{ResNet-18} & \multirow{2}*{11.17M} & \xmark & 76.4 & 0.0 & 8.3 & 0.2 & 37.3 & 0.0 \\
     & &  & \checkmark & 54.2 & 27.8 & 30.7 & 29.7 & 49.9 & 24.0 \\ 
     \cmidrule{2-10}
    & \multirow{2}*{WideResNet32} & \multirow{2}*{46.16M} & \xmark & 80.2 & 0.0 & 17.1 & 1.2 & 41.6 & 0.0  \\
    &  &  & \checkmark & 59.6 & 30.9 & 34.1 & 32.7 & 55.1 & \textbf{27.2} \\
     \cmidrule{2-10}
    & IDC(Ours) & 42.84M & \xmark & 59.4 & \textbf{33.4} & \textbf{42.0} & \textbf{38.8} & \textbf{56.0} & 17.7 \\
    \bottomrule
  \end{tabular}
    \caption{The adversarial robustness evaluation of CNN-based models and our method on both CIFAR-10 and CIFAR-100. 'AT' denotes the adversarial training for the networks.}
    \label{tab:cifar10that}
\end{table*}

\begin{table}
    \centering
\resizebox{0.49\textwidth}{!}{
    \scriptsize
  \begin{tabular}{c|c|c|c|c|c|c}
    \toprule
    Method & AT & Natural & $\text{PGD}^{20}$ & FGSM & CW & AutoAttack \\
    \midrule
    \multirow{2}*{ResNet-18} & \checkmark & 38.06 & 20.21 & 21.44 & 32.52 & 16.91 \\ 
      & \xmark & 44.40 & 0.26 & 5.14 & 12.23 & 0.00 \\
     \midrule
   \multirow{2}*{WideResNet32} & \checkmark & 44.63 & \textbf{24.21} & \textbf{25.88} & \textbf{38.41} & \textbf{20.93} \\
     & \xmark & 46.80 & 0.01 & 0.84 & 12.95 & 0.0 \\
     \midrule
    IDC(Ours) & \xmark & 37.65 & 16.03 & 22.84 & 34.33 & 13.48$\dagger$ \\
    \bottomrule
  \end{tabular}
  }
    \caption{The adversarial robustness comparison with CNN-based defense methods on Tiny-ImageNet.}
    \label{tab:tinythat}
\end{table}

\subsection{Comparison with CNN-based Methods}

\subsubsection{CIFAR-10.}

The upper part of Table \ref{tab:cifar10that} shows the performance of the CNN-based defense models and our IDC on CIFAR-10 dataset against $\text{PGD}^{20}$, FGSM, MIFGSM, CW and AutoAttack $\ell_{\infty}$ ($\epsilon=8/255$) where $\text{PGD}^{20}$ denotes the attack steps of PGD is 20. IDC maintains an accuracy exceeding 80\% on $\text{PGD}^{20}$, MIFGSM, and 59.9\% on AutoAttack while CNN-based methods without AT perform near zero accuracy. Even compared with AT methods, IDC improves an average robust accuracy by 20.9\% over ResNet-18 and 17.18\% over WideResNet32. Among them, our IDC improves the robust accuracy by 32.6\% and 29.7\% against $\text{PGD}^{20}$ compared with ResNet-18 and WideResNet32.

\subsubsection{CIFAR-100.}

We compare the robust performance on CIFAR-100 against $\text{PGD}^{20}$, FGSM, MIFGSM, CW and AutoAttack $\ell_{\infty}$ ($\epsilon=8/255$) on the lower part of Table \ref{tab:cifar10that}. Our IDC maintains competitive standard accuracy with CNN-based classifiers with AT. As for the robust accuracy, the IDC can improve by 4.7\%, 10.7\%, 8.5\% over ResNet-18 and 1.7\%, 6.6\%, 4.7\% over WideResNet32 against $\text{PGD}^{20}$, FGSM, MIFGSM attack. It also shows competitive performance against CW attack and AutoAttack. 

\begin{table}
    \centering
\resizebox{0.48\textwidth}{!}{
    \scriptsize
  \begin{tabular}{c|c|c|c|c}
    \toprule
    \multirow{2}*{Method} &  \multirow{2}*{Params}  & \multirow{2}*{Standard} & \multicolumn{2}{c}{Robust Acc} \\
      &  &   & BPDA+EOT & PGD+EOT \\
    \midrule
    ResNet-18 & 11.17M  & 82.65 & 51.08 & 51.17 \\
    WideResNet32 & 46.16M & 86.59 & 54.24 & 54.25 \\
    \midrule
    RDC$\dagger$ & 111.48M  & 93.16 & 73.24 & - \\
    MimicDiffusion$\dagger$ & 140.4M & 92.50 & \textbf{92.00} & - \\
    ContrastDiffPure$\dagger$ & 278M & 92.61 & 81.94 & - \\
    DiffPure$\dagger$ & 244.30M  & 89.02 & 81.40 & 46.84 \\
    \midrule
    IDC(Ours) & 9.39M  & 86.10 & 82.97 & \textbf{84.70}  \\
    \bottomrule
  \end{tabular}
  }
    \caption{The adversarial robustness evaluation of both CNN-based and DM-based methods against BPDA+EOT and PGD+EOT attack on CIFAR-10. The symbol $\dagger$ denotes the evaluation is on a subset of the dataset.}
    \label{tab:cifar10bpda}
\end{table}

\subsubsection{Tiny-ImageNet.}

Compared to AT methods, IDC performs better than ResNet-18 and slightly less than WideResNet32. However, compared to methods without AT, IDC has significant performance advantages. We will consider improving the robustness of IDC in complex datasets in the future.

\begin{figure*}
    \centering
    \subfigure[Robustness under different PGD attack steps]{
        \includegraphics[width=0.343\linewidth]{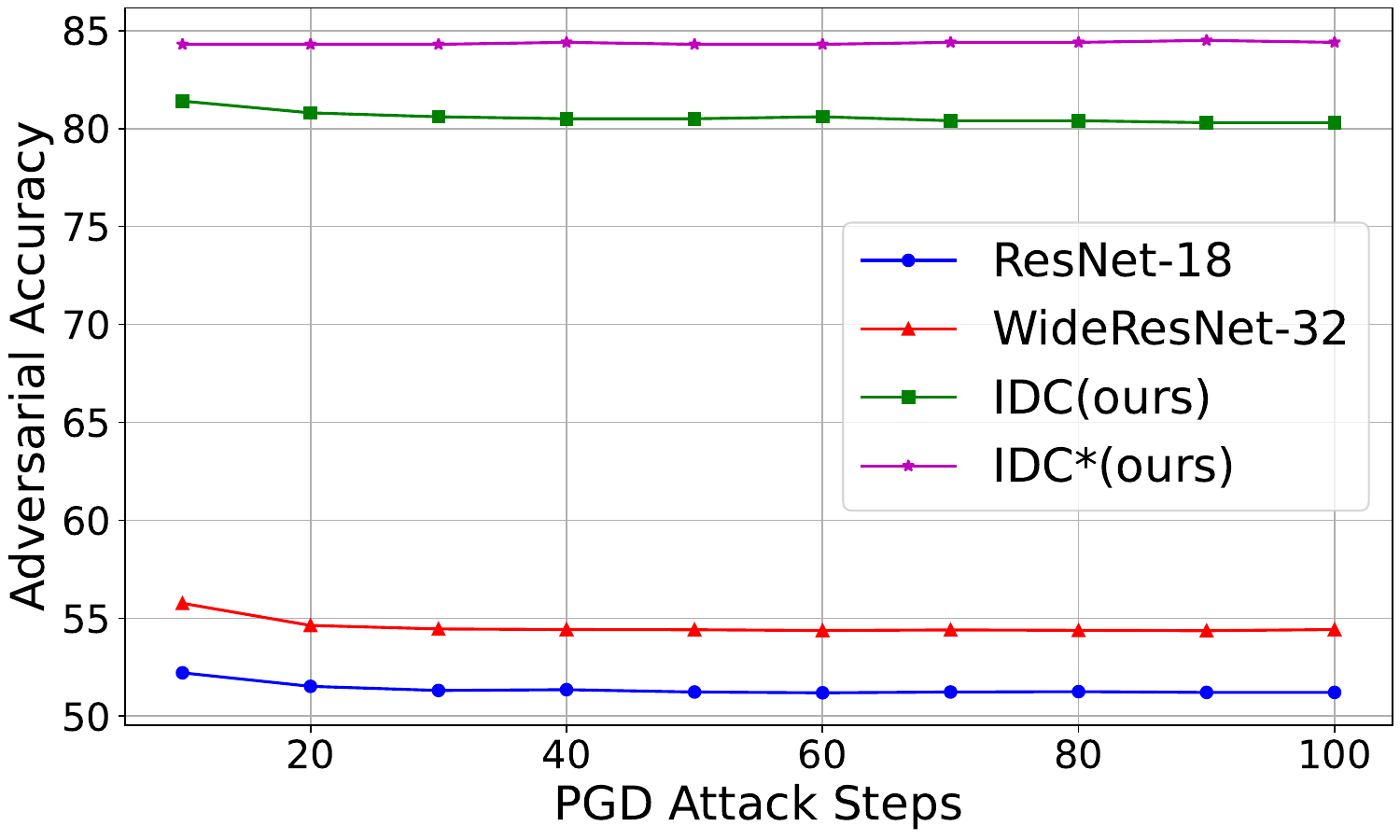}
        \label{f-pgdk}
    }
    \subfigure[Robustness under different PGD attack eps]{	   \includegraphics[width=0.343\linewidth]{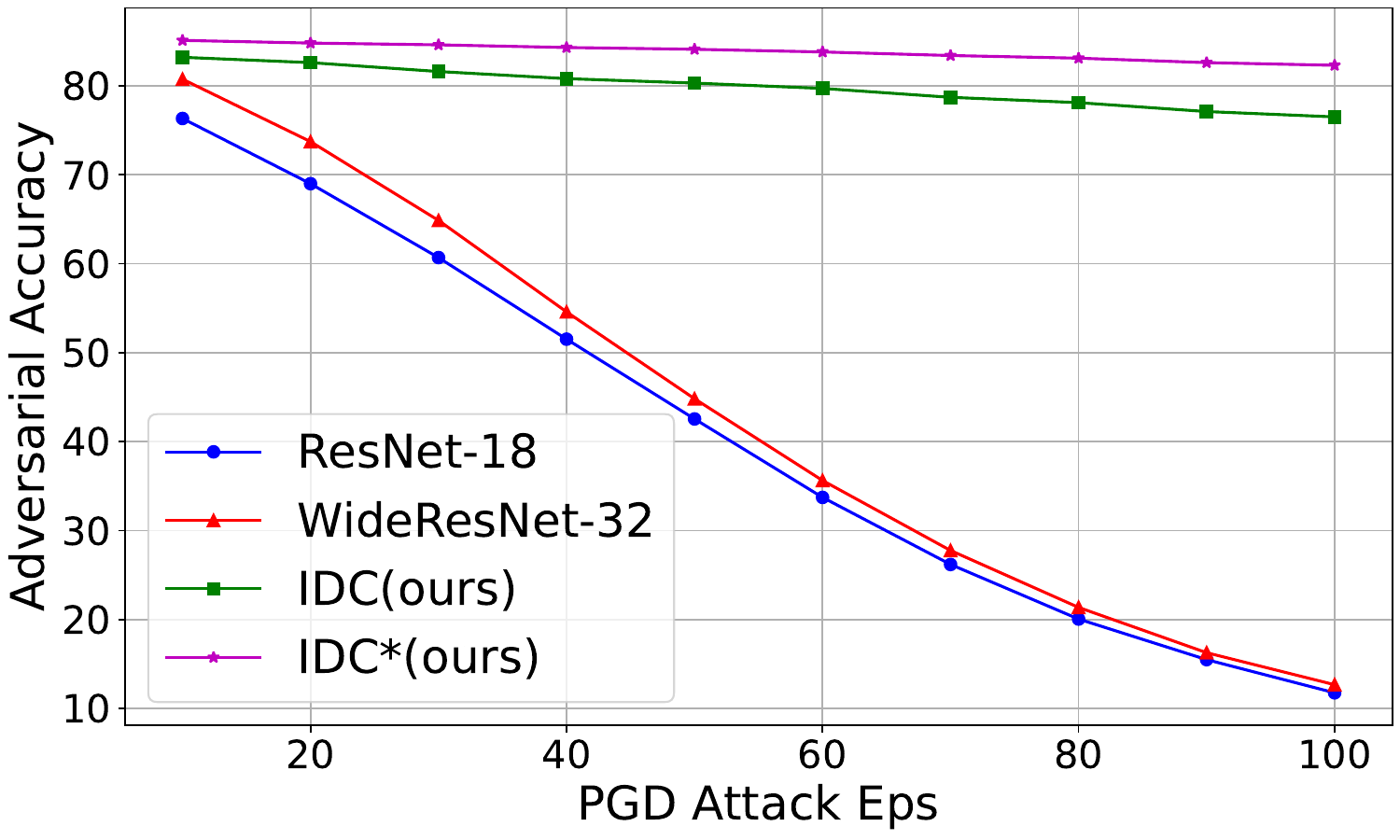}
        \label{f-pgdeps}
    }
    \subfigure[Ablation of diffusion timesteps]{	   \includegraphics[width=0.27\linewidth]{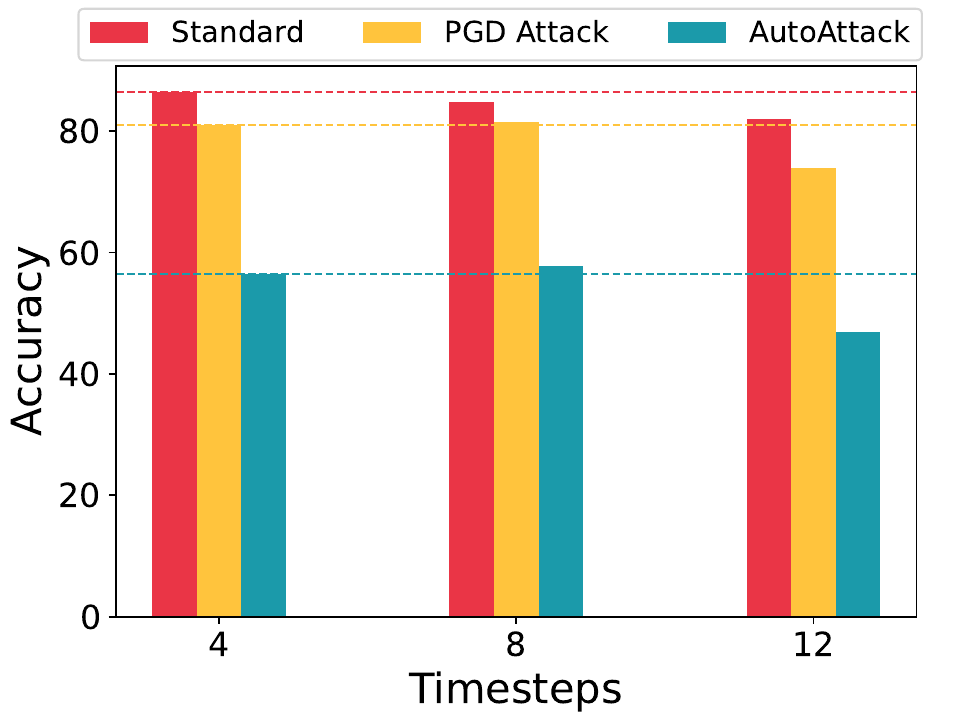}
        \label{f-timestep}
    } 
    \caption{The evaluation of robustness under different PGD attack settings and different diffusion timesteps. (a) denotes larger attack iterations. (b) denotes larger perturbation sizes. The symbol * denotes the adversarial robustness of IDC without inter-class loss. (c) Standard and robust accuracy under PGD attack and AutoAttack $\ell_{\infty}$ ($\epsilon=8/255$) with different timesteps.}
    \label{f-pgdablation}
\end{figure*}

\subsection{Defense Against Adaptive Attacks}

\subsubsection{CIFAR-10.}
We compare our method against BPDA+EOT and PGD+EOT attacks in Table \ref{tab:cifar10bpda}. Different from other DM-based defense methods, we conduct a full evaluation of CIFAR-10. Compared with ResNet-18 and WideResNet32, IDC improves the robust accuracy by 31.89\%, 28.73\% on BPDA+EOT attack and 33.53\%, 30.45\% on PGD+EOT attack with lower parameters.
As for DM-based methods, IDC achieves more competitive performance with significantly reduced computations. 
Concretely, compared with DiffPure~\cite{nie2022diffusion}, we improve the accuracy by 1.57\% against BPDA+EOT attacks and 37.86\% against PGD+EOT attacks, with only 3.84\% of the parameters. Compared with novel approaches RDC, MimicDiffusion~\cite{song2024mimicdiffusion} and ContrastDiffPure~\cite{bai2024diffusion}, IDC also shows competitive performance with exceptionally lower parameters.

\subsubsection{CIFAR-100.} 

Table \ref{tab:cifar100bpda} shows the robustness performance against BPDA+EOT attack on CIFAR-100 while our method also achieves the competitive standard accuracy with fewer parameters. The robust accuracy of IDC is also advanced with the improvement over the ResNet-18 by 0.49\%.

\subsection{Ablation Study}

\subsubsection{Effectiveness of Proposed Modules.} We ablate the proposed orthogonal image labels and inter-class loss in Table \ref{tab:cifar10orthogonal}. The inter-class loss can improve the robust accuracy by 3.8\% against PGD attack and by 3.4\% against AutoAttack. Incorporating inter-class loss can also consistently enhance the adversarial robustness of the PGD attack in Figure \ref{f-pgdablation}.
The above results demonstrate that using the inter-class loss can effectively enhance the ability to generate images that are distinguishable among different class labels, thereby strengthening the resilience against attacks. For the orthogonal image labels, to maintain similar distances between image labels like traditional one-hot labels, we enhance the orthogonality between images. Table \ref{tab:cifar10orthogonal} confirms the efficacy of this configuration, while orthogonal image labels enhance the network's robustness performance by 21.1\% under the PGD attack and by 25.3\% under AutoAttack.

\begin{table}
    \centering
\resizebox{0.48\textwidth}{!}{
    \scriptsize
  \begin{tabular}{c|c|c|c|c}
    \toprule
    Method &  Params & AT & Standard Acc & Robust Acc \\
    \midrule
    ResNet-18 & 11.17M & \checkmark & 54.23 & 27.53  \\
    WideResNet32 & 46.16M  & \checkmark  & 59.57 & \textbf{30.36} \\
     \midrule
    IDC(Ours) & 42.84M  & \xmark & 59.97 & 28.02 \\
    \bottomrule
  \end{tabular}
  }
    \caption{Comparison with CNN-based defense methods using the BPDA+EOT attack on CIFAR-100.}
    \label{tab:cifar100bpda}
\end{table}

\begin{table}
    \centering
  \begin{tabular}{c|c|c|c|c}
    \toprule
     \multirow{2}*{Inter-C}&\multirow{2}*{OL} & \multirow{2}*{Standard Acc} & \multicolumn{2}{c}{Robust Acc} \\
      &  &  & $\text{PGD}^{20}$ & AutoAttack \\
    \midrule
    \checkmark & \xmark & 87.5 & 63.7 & 34.6 \\
     \xmark & \checkmark & 86.4 & 81.0  & 56.5\\
    \checkmark & \checkmark & 86.0 & \textbf{84.8} & \textbf{59.9} \\
    
    \bottomrule
  \end{tabular}
    \caption{Effectiveness of orthogonal image labels in our methods with PGD attack and AutoAttack $\ell_{\infty}$ ($\epsilon=8/255$) on CIFAR-10. 'Inter-C' denotes incorporating the inter-class loss, while 'OL' denotes the orthogonality of labels setting.}
    \label{tab:cifar10orthogonal}
\end{table}

\subsubsection{Different PGD Attacks Settings.}

We adjust the steps and $\epsilon$ of the PGD attack, and Figure \ref{f-pgdablation} shows the robustness performance of our IDC and CNN-based methods. As shown in Figure \ref{f-pgdk}, the IDC can maintain robustness performance as the attack steps increase, achieving a performance improvement of 32.99\% and 29.8\% compared to ResNet-18 and WideResNet, respectively. In Figure \ref{f-pgdeps}, as the attack $\epsilon$ increases, CNN-based methods experience a significant performance decline, whereas our method manages to maintain relatively stable performance. Compared to WideResNet32, our method shows an improvement of 4.36\% at $2/255$, and this improvement escalates to 69.64\% when $\epsilon=20/255$.

\subsubsection{Different Diffusion Timesteps.}

For a better illustration of the reduction of diffusion timesteps, we compile statistics on the standard accuracy and robustness accuracy under PGD and AutoAttack of the IDC in Figure \ref{f-timestep}, with timesteps $L_s$ of 4, 8, and 12, respectively. The performance at timesteps $L_s=8$ is comparable to that at $L_s=4$, but it incurs a substantial increase in computational cost during inference. When $L_s=12$, the standard performance decreases by 4.5\%, and the adversarial robustness declines by 7.1\% and 9.6\%, respectively. A possible reason is that when there are excessive timesteps, the noise addition process in each timestep is not sufficiently pronounced, inadvertently increasing the learning difficulty for the network.

\section{Conclusion}

In this paper, we propose a new framework called IDC for DM-based adversarial robustness, which transforms the task of DMs from generating high-quality images to predicting distinguishable image labels. Based on our framework, we streamline the diffusion network, evaluate performance across the entire dataset, and compare it with both CNN-based and DM-based methods. A classification loss is also incorporated to enhance the adaptability of the DM-based framework to adversarial robustness. We select multiple representative attacks on the CIFAR-10 and CIFAR-100 datasets to demonstrate that our IDC achieves competitive results in terms of parameters and adversarial robustness.

\bibliography{aaai25}

\clearpage
\appendix

\section{A. Specified Derivation}

\subsection{A.1 Derivation of forward transition probability}

During the forward diffusion of the Brownian Bridge process, the transition from initial state $\boldsymbol{y}_0$ to the state of timestep $t-1$ and $t$ can be formulated as:
\begin{equation}
    \boldsymbol{y}_t = (1-m_t)\boldsymbol{y}_0+m_t\boldsymbol{x}+\sqrt{\delta_t}\boldsymbol{\epsilon}_t,
    \label{y0_to_yt_2}
\end{equation}
\begin{equation}
    \boldsymbol{y}_{t-1} = (1-m_{t-1})\boldsymbol{y}_0+m_{t-1}\boldsymbol{x}+\sqrt{\delta_{t-1}}\boldsymbol{\epsilon}_{t-1}.
    \label{y0_to_yt-1}
\end{equation}
By solving Eq. (\ref{y0_to_yt-1}), the expression of $\boldsymbol{y}_0$ with respect to $\boldsymbol{y}_{t-1}$ is obtained as follows:
\begin{equation}
    \boldsymbol{y}_{0} = \frac{\boldsymbol{y}_{t-1}-m_{t-1}\boldsymbol{x}-\sqrt{\delta_{t-1}}\boldsymbol{\epsilon}_{t-1}}{1-m_{t-1}}.
    \label{yt-1_to_y0}
\end{equation}
Next, we substitute Eq. (\ref{yt-1_to_y0}) into Eq. (\ref{y0_to_yt_2}) and formulate the forward process as:
\begin{equation}
\begin{aligned}
    \boldsymbol{y}_{t} &= (1-m_t)\frac{\boldsymbol{y}_{t-1}-m_{t-1}\boldsymbol{x}-\sqrt{\delta_{t-1}}\boldsymbol{\epsilon}_{t-1}}{1-m_{t-1}} \\
    &+m_t\boldsymbol{x}+\sqrt{\delta_t}\boldsymbol{\epsilon}_t \\
    &=\frac{1-m_t}{1-m_{t-1}}\boldsymbol{y}_{t-1} + (m_t - \frac{1-m_t}{1-m_{t-1}}m_{t-1})\boldsymbol{x} \\
    &+\sqrt{\delta_t}\boldsymbol{\epsilon}_t - \frac{1-m_t}{1-m_{t-1}}\sqrt{\delta_{t-1}}\boldsymbol{\epsilon}_{t-1} \nonumber
    \label{yt-1_to_yt}.
\end{aligned}
\end{equation}
For simplicity, we set $\gamma=\frac{1-m_t}{1-m_{t-1}}$. In the forward process of transition $q(\boldsymbol{y}_t\mid \boldsymbol{y}_{t-1}, \boldsymbol{x})$, as $\boldsymbol{\epsilon}_t, \boldsymbol{\epsilon}_{t-1} \sim \mathcal{N}(\boldsymbol{0}, \boldsymbol{I})$, the $\mu$ of $q(\boldsymbol{y}_t\mid \boldsymbol{y}_{t-1}, \boldsymbol{x})$ is 
$\gamma \boldsymbol{y}_{t-1}+(m_t-\gamma m_{t-1})\boldsymbol{x}$.
Based on the definition of conditional variance~\cite{bollerslev1986generalized}, we need to consider the additional variation $\delta_{t\mid t-1}$ in $\boldsymbol{y}_{t}$ given $\boldsymbol{y}_{t-1}$ and $\boldsymbol{x}$, which can be formulated as
$\delta_{t\mid t-1} = \delta_{t} - \gamma^2 \delta_{t-1}$.
Finally, the forward process of transition $q(\boldsymbol{y}_t\mid \boldsymbol{y}_{t-1}, \boldsymbol{x})$ can be defined as:
\begin{equation}
\begin{aligned}
q(\boldsymbol{y}_t\mid \boldsymbol{y}_{t-1}, \boldsymbol{x}) = \mathcal{N}&(\boldsymbol{y}_t; \gamma \boldsymbol{y}_{t-1}+(m_t-\gamma m_{t-1})\boldsymbol{x}, \\
&(\delta_{t} - \gamma^2 \delta_{t-1})\boldsymbol{I}). \nonumber
\end{aligned}
\end{equation}

\subsection{A.2 Derivation of the objective of intra-class loss}

For timestep $t$, the reverse process is $q(\boldsymbol{y}^i_{t-1}\mid \boldsymbol{y}^i_t, \boldsymbol{y}^i_0, \boldsymbol{x})=\mathcal{N}(\boldsymbol{y}^i_{t-1};\mu(\boldsymbol{y}^i_{t}, \boldsymbol{x}), \delta \boldsymbol{I})$ where the mean value $\mu$ is the noise predictor, and the prediction of the DM can be expressed as $p_\theta(\hat{\boldsymbol{y}}^i_{t-1}\mid \boldsymbol{y}^i_t, \boldsymbol{x})=\mathcal{N}(\hat{\boldsymbol{y}}^i_{t-1}; \hat{\mu}_\theta(\boldsymbol{y}^i_t, \boldsymbol{x}, t), \delta_t \boldsymbol{I})$ where $\theta$ is the parameter of the network optimized by $\mu_\theta$. The Evidence Lower Bound (ELBO) for optimization of $\mathcal{L}_{intra}$ can be formulated as:
\begin{equation}
\displaystyle
\scalebox{0.92}{$
\begin{aligned}
&- \mathbb{E}_q \Big[\sum_{t=2}^{T_s} D_{KL}(q(\boldsymbol{y}^i_{t-1}\mid \boldsymbol{y}^i_t, \boldsymbol{y}^i_0, \boldsymbol{x})\Vert p_\theta(\hat{\boldsymbol{y}}^i_{t-1}\mid \boldsymbol{y}^i_t, \boldsymbol{x})) \\
&- \log p_\theta(\hat{\boldsymbol{y}}^i_0\mid \boldsymbol{y}^i_1, \boldsymbol{x}) + D_{KL}(q(\boldsymbol{y}^i_{T}\mid \boldsymbol{y}^i_0, \boldsymbol{x})\Vert p(\boldsymbol{y}^i_{T}\mid \boldsymbol{x}) \Big].
\label{ELBO_recon}
\end{aligned}
$}
\end{equation}
As $\boldsymbol{y}^i_{T}$ is equal to $\boldsymbol{x}$, the last term of Eq.\ref{ELBO_recon} can be ignored. We aim to simplify the training objective based on the noise predictor $\mu(\boldsymbol{y}^i_{t}, \boldsymbol{y}^i_{0}, \boldsymbol{x})$. Following the derivation in BBDM~\cite{li2023bbdm}, the reverse process of transition $q(\boldsymbol{y}^i_{t-1}\mid \boldsymbol{y}^i_t, \boldsymbol{y}^i_0, \boldsymbol{x})=\mathcal{N}(\boldsymbol{y}^i_{t-1};\mu(\boldsymbol{y}^i_{t}, \boldsymbol{y}^i_{0}, \boldsymbol{x}), \delta \boldsymbol{I})$, where the mean value and the variance can be derived as follows:
\begin{equation}
\begin{aligned}
\mu\left(\boldsymbol{y}^i_t, \boldsymbol{y}^i_0, \boldsymbol{x}\right) & =\frac{\delta_{t-1}}{\delta_t} \gamma \boldsymbol{y}^i_t +(1-m_{t-1}) \frac{\delta_{t \mid t-1}}{\delta_t} \boldsymbol{y}^i_0 \\
& +(m_{t-1}-m_t \gamma \frac{\delta_{t-1}}{\delta_t}) \boldsymbol{x},
\label{mu_recon}
\end{aligned}
\end{equation}
\begin{equation}
\delta=\frac{\delta_{t \mid t-1} \cdot \delta_{t-1}}{\delta_t}. \nonumber
\end{equation}
Considering the $\boldsymbol{y}^i_0$ is the image labels during training and is unknown during inference, we derive its distribution as the objective of $\epsilon$ through the reparametrization method~\cite{ho2020denoising}. Based on the Eq. (\ref{y0_to_yt_2}), the expression of $\boldsymbol{y}^i_0$ can be formulated as:
\begin{equation}
\boldsymbol{y}^i_0 = \boldsymbol{y}^i_t - m_t(\boldsymbol{x} - \boldsymbol{y}^i_0) - \sqrt{\delta_t}\boldsymbol{\epsilon}_t.
\label{x_yt_to_y0}
\end{equation}
By substituting Eq. (\ref{x_yt_to_y0}) into Eq. (\ref{mu_recon}), the mean value can be reformulated as follows:
\begin{equation}
\begin{aligned}
\mu\left(\boldsymbol{y}^i_t, \boldsymbol{x}\right) & =\frac{\delta_{t-1}}{\delta_t} \gamma \boldsymbol{y}^i_t + (1-m_{t-1}) \frac{\delta_{t \mid t-1}}{\delta_t}[\boldsymbol{y}^i_t \\
&-m_t(\boldsymbol{x}-\boldsymbol{y}^i_0)-\sqrt{\delta_t}\epsilon_t] \\
&+(m_{t-1}-m_t \gamma \frac{\delta_{t-1}}{\delta_t}) \boldsymbol{x} \\
&=(m_{t-1}-m_t \gamma \frac{\delta_{t-1}}{\delta_t})\boldsymbol{x} \\
&+[\frac{\delta_{t-1}}{\delta_t} \gamma+(1-m_{t-1})\frac{\delta_{t \mid t-1}}{\delta_t}]\boldsymbol{y}^i_t \\
&-[(1-m_{t-1})\frac{\delta_{t \mid t-1}}{\delta_t}](m_t(\boldsymbol{x}-\boldsymbol{y}^i_0)+\sqrt{\delta_t}\epsilon_t). \nonumber
\end{aligned}
\end{equation}
We extract the coefficients as:
\begin{equation}
\begin{aligned}
& c_{x t}=m_{t-1}-m_t \gamma \frac{\delta_{t-1}}{\delta_t}, \\
& c_{y t}=\frac{\delta_{t-1}}{\delta_t} \gamma+\left(1-m_{t-1}\right)\frac{\delta_{t \mid t-1}}{\delta_t}, \\
& c_{\epsilon t}=-\left(1-m_{t-1}\right) \frac{\delta_{t \mid t-1}}{\delta_t}. \nonumber
\end{aligned}
\end{equation}
The mean value can ultimately be simplified to the form expressed in the paper as $\mu(\boldsymbol{y}^i_t, \boldsymbol{x})=c_{x t} \boldsymbol{x}+c_{y t} \boldsymbol{y}^i_t+c_{\epsilon t} (m_t(\boldsymbol{x}-\boldsymbol{y}^i_0) + \sqrt{\delta_t}\boldsymbol{\epsilon}_t)$. To optimize $\epsilon$, we reformulate the mean value of prediction $p_\theta(\hat{\boldsymbol{y}}^i_{t-1}\mid \boldsymbol{y}^i_t, \boldsymbol{x})$ as $\hat{\mu}_{\theta}(\boldsymbol{y}^i_t, \boldsymbol{x}, t)=c_{x t} \boldsymbol{x}+c_{y t} \boldsymbol{y}^i_t+c_{\epsilon t} \epsilon_\theta(\boldsymbol{y}^i_t, t)$. Thus, the training objective loss can be formulated as:
\begin{equation}
\mathcal{L}_{intra} = \Vert m_t(\boldsymbol{x}-\boldsymbol{y}^i_0) + \sqrt{\delta_t}\boldsymbol{\epsilon}_t -\epsilon_\theta(\boldsymbol{y}^i_t, t) \Vert_1. \nonumber
\end{equation}

\subsection{A.3 Derivation of the prediction}

With the significant reduction in timesteps, we maintain the inference timesteps the same as training timestep $T_s$. Given the diffusion timestep $T_s$ and the input sample $\boldsymbol{x}=\boldsymbol{y}_T$, the inference process follows the reverse process $p_\theta(\hat{\boldsymbol{y}}_{t-1}\mid \boldsymbol{y}_t, \boldsymbol{x})=\mathcal{N}(\hat{\boldsymbol{y}}_{t-1} ; \hat{\mu}_\theta(\boldsymbol{y}_t, \boldsymbol{x}, t), \delta \boldsymbol{I})$ to generate sample
distribution. Based on the noise predictor formulation $\mu(\boldsymbol{y}_t, \boldsymbol{x})=c_{x t} \boldsymbol{x}+c_{y t} \boldsymbol{y}_t+c_{\epsilon t} (m_t(\boldsymbol{x}-\boldsymbol{y}_0) + \sqrt{\delta_t}\boldsymbol{\epsilon}_t)$ and the reformulation of $\hat{\mu}_{\theta}(\boldsymbol{y}_t, \boldsymbol{x}, t)=c_{x t} \boldsymbol{x}+c_{y t} \boldsymbol{y}_t+c_{\epsilon t} \epsilon_\theta(\boldsymbol{y}_t, t)$, the sampling at $t<T_s$ can be formulated as:
\begin{equation}
\hat{\boldsymbol{y}}_{t-1} = c_{x t} \boldsymbol{x}+c_{y t} \boldsymbol{y}_t + c_{\epsilon t} \epsilon_\theta(\boldsymbol{y}_t, t) + \sqrt{\delta}\boldsymbol{\epsilon}. \nonumber
\end{equation}
At $t=0$, we sample the final generated image without random noise, which can be formulated as:
\begin{equation}
\hat{\boldsymbol{y}}_{0} = c_{x 1} \boldsymbol{x}+c_{y 1} \boldsymbol{y}_1 + c_{\epsilon 1} \epsilon_\theta(\boldsymbol{y}_1, 1).
\label{pre_y0}
\end{equation}
The predicted class $c$ by IDC can be formulated through $L_1$ distance of the designed image labels and generated images in Eq. (\ref{pre_y0}) as:
\begin{equation}
c = \mathop{\arg\min}\limits_{i} \Vert \hat{\boldsymbol{y}}_0 - \boldsymbol{y}^i_0 \Vert_1,
\label{pre_y_2}
\end{equation}
where $\boldsymbol{y}^i_0$ denotes the pre-defined image label of class $i$. During the evaluation of adversarial robustness, to maintain the differentiability of the network, we reformulate the prediction Eq. (\ref{pre_y_2}) as:
\begin{equation}
logits = \frac{e^{-\tau d_i}}{\sum^K_{i=0}e^{-\tau d_i}}, i = \{0, 1, ..., K\}, \nonumber
\end{equation}
where $\tau$ is a hyper-parameter and is set to 0.1 in this paper, and the distance between the generated images and different image labels is $d_i = \Vert \hat{\boldsymbol{y}}_0 - \boldsymbol{y}^i_0 \Vert_1$.

\section{B. Proofs}

\subsection{B.1 Proofs of the proposition 1}

In the objective of loss $\mathcal{L}_{inter}$, we consider the most confusing class $j$ of image $x$ where the image label $\boldsymbol{y}^j_0$ is the most closest one to the translated $\hat{\boldsymbol{y}}^i_0$, as $j=\argmin_j \Vert \hat{\boldsymbol{y}}^i_0 - \boldsymbol{y}^j_0 \Vert_1$ and $j \neq i$. During the optimization, the distance between $\hat{\boldsymbol{y}}^i_0$ and the most confusing image label $\boldsymbol{y}^j_0$ is expected to be maximized, which can be formulated in a reverse format of ELBO as:
\begin{equation}
\displaystyle
\scalebox{0.95}{$ - \mathbb{E}_q \Big[-\sum_{t=2}^{T_s} D_{KL}(q(\boldsymbol{y}^j_{t-1}\mid \boldsymbol{y}^j_t, \boldsymbol{y}^j_0, \boldsymbol{x})\Vert p_\theta(\hat{\boldsymbol{y}}^i_{t-1}\mid \boldsymbol{y}^i_t, \boldsymbol{x})) \Big].
$}
\label{ELBO_inter_2}
\end{equation}
Similar to the deviation of the reverse process $q(\boldsymbol{y}_{t-1}\mid \boldsymbol{y}_t, \boldsymbol{y}_0, \boldsymbol{x})$ in Section A.2, the reverse process can be formulated as $q(\boldsymbol{y}^j_{t-1}\mid \boldsymbol{y}^j_t, \boldsymbol{y}^j_0, \boldsymbol{x})=\mathcal{N}(\boldsymbol{y}^j_{t-1};\mu(\boldsymbol{y}^j_{t}, \boldsymbol{y}^j_{0}, \boldsymbol{x}), \delta \boldsymbol{I})$ where the noise predictor $\mu(\boldsymbol{y}^j_{t}, \boldsymbol{y}^j_{0}, \boldsymbol{x})$ is simplifid to the function of $\boldsymbol{\epsilon}$ through the reparametrization method~\cite{ho2020denoising} and formulated as:
\begin{equation}
\mu(\boldsymbol{y}^j_t, \boldsymbol{x})=c_{x t} \boldsymbol{x}+c_{y t} \boldsymbol{y}^j_t+c_{\epsilon t} (m_t(\boldsymbol{x}-\boldsymbol{y}^j_0) + \sqrt{\delta_t}\boldsymbol{\epsilon}_t),
\label{mu_ytj}
\end{equation}
where the coefficients are:
\begin{equation}
\begin{aligned}
& c_{x t}=m_{t-1}-m_t \gamma \frac{\delta_{t-1}}{\delta_t}, \\
& c_{y t}=\frac{\delta_{t-1}}{\delta_t} \gamma+\left(1-m_{t-1}\right)\frac{\delta_{t \mid t-1}}{\delta_t}, \\
& c_{\epsilon t}=-\left(1-m_{t-1}\right) \frac{\delta_{t \mid t-1}}{\delta_t}.
\end{aligned}
\label{param}
\end{equation}
When the objective is transferred to the noise $\boldsymbol{\epsilon}$, the mean value of prediction distribution $p_\theta(\hat{\boldsymbol{y}}^i_{t-1}\mid \boldsymbol{y}^i_t, \boldsymbol{x})$ needs to be reformulated as the following form, which shares the same coefficients and mean values with Eq. (\ref{mu_ytj}).
\begin{equation}
\hat{\mu}_{\theta}(\boldsymbol{y}^i_t, \boldsymbol{x}, t)=c_{x t} \boldsymbol{x}+c_{y t} \boldsymbol{y}^j_t+c_{\epsilon t} \epsilon_\theta(\boldsymbol{y}^i_t, t).
\label{pre_mu_j}
\end{equation}
However, during the simplification of intra-class loss in Section A.2, the mean value of $p_\theta(\hat{\boldsymbol{y}}^i_{t-1}\mid \boldsymbol{y}^i_t, \boldsymbol{x})$ has been reformulated to the form as:
\begin{equation}
\hat{\mu}_{\theta}(\boldsymbol{y}^i_t, \boldsymbol{x}, t)=c_{x t} \boldsymbol{x}+c_{y t} \boldsymbol{y}^i_t+c_{\epsilon t} \epsilon_\theta(\boldsymbol{y}^i_t, t).
\label{pre_mu_i}
\end{equation}
Compared the form of Eq. (\ref{pre_mu_j}) and Eq. (\ref{pre_mu_i}), there is a gap between $\boldsymbol{y}^i_t$ and $\boldsymbol{y}^j_t$, causing by the ground truth $\mu$ of different labels $i$ and $j$. To bridge this gap, we derive a single timestep process from $\boldsymbol{y}^i_{t-1}$ to $\boldsymbol{y}^j_t$ and reformulate the forward process in Eq. (\ref{y0_to_yt_2}) as:
\begin{equation}
\begin{aligned}
\boldsymbol{y}^{js}_t = &\boldsymbol{y}^i_{t-1} + \boldsymbol{y}^j_t - \boldsymbol{y}^j_{t-1} \\
= &(1-m_{t-1})\boldsymbol{y}^i_{0} + m_{t-1} \boldsymbol{x} + \sqrt{\delta_{t-1}}\boldsymbol{\epsilon}_{t-1} \\
&+ (1-m_{t})\boldsymbol{y}^j_{0} + m_{t} \boldsymbol{x} + \sqrt{\delta_{t}}\boldsymbol{\epsilon}_t \\
&- (1-m_{t-1})\boldsymbol{y}^j_{0} - m_{t-1} \boldsymbol{x} - \sqrt{\delta_{t-1}}\boldsymbol{\epsilon}_{t-1} \\
=&(1-m_{t-1})\boldsymbol{y}^i_{0} + (m_{t-1}-m_{t})\boldsymbol{y}^j_{0}+m_t\boldsymbol{x} + \sqrt{\delta_{t}}\boldsymbol{\epsilon}_t. 
\end{aligned}
\label{y0_to_ytjs}
\end{equation}
Here we approximate the $\boldsymbol{y}_t$ in the Eq. (\ref{y0_to_yt_2}) to narrow the difference between $\boldsymbol{y}^{i}_t$ and $\boldsymbol{y}^{j}_t$ to a single timestep, thus we can keep the formulation of $\hat{\mu}_{\theta}(\boldsymbol{y}^i_t, \boldsymbol{x}, t)$ in the Section A.2 and applies to the noise predictor of both $\mu(\boldsymbol{y}^i_t, \boldsymbol{x})$ in intra-class loss and $\mu(\boldsymbol{y}^j_t, \boldsymbol{x})$ in inter-class loss. Next, we need to derive the new forms of noise supervision based on the process in Eq. (\ref{y0_to_ytjs}) and coefficients in Eq. (\ref{param}). According to the reparametrization method~\cite{ho2020denoising}, we can reformulated $\boldsymbol{y}_0$ in Eq. (\ref{x_yt_to_y0}) as follows:
\begin{equation}
\displaystyle
\scalebox{0.96}{$
\boldsymbol{y}_0 = \boldsymbol{y}^{js}_t - m_t(\boldsymbol{x}-\boldsymbol{y}^j_0)-m_{t-1}(\boldsymbol{y}^j_0-\boldsymbol{y}^i_0)-\sqrt{\delta_{t}}\boldsymbol{\epsilon}_t
$}.
\label{x_ytj_to_y0}
\end{equation}
By substituting Eq. (\ref{x_ytj_to_y0}) into Eq. (\ref{mu_recon}), the mean value can be reformulated as follows:
\begin{equation}
\begin{aligned}
\mu\left(\boldsymbol{y}^{js}_t, \boldsymbol{x}\right) &=\frac{\delta_{t-1}}{\delta_t} \gamma \boldsymbol{y}^{js}_t + (1-m_{t-1}) \frac{\delta_{t \mid t-1}}{\delta_t}[\boldsymbol{y}^{js}_t \\
&-m_t(\boldsymbol{x}-\boldsymbol{y}^j_0)-m_{t-1}(\boldsymbol{y}^j_0-\boldsymbol{y}^i_0)-\sqrt{\delta_{t}}\boldsymbol{\epsilon}_t] \\
&+(m_{t-1}-m_t \gamma \frac{\delta_{t-1}}{\delta_t}) \boldsymbol{x} \\
&=(m_{t - 1} - m_t \gamma \frac{\delta_{t-1}}{\delta_t})\boldsymbol{x} \\
&+[\frac{\delta_{t-1}}{\delta_t} \gamma +(1-m_{t-1})\frac{\delta_{t \mid t-1}}{\delta_t}]\boldsymbol{y}^{js}_t \\
&-[(1 - m_{t-1})\frac{\delta_{t \mid t-1}}{\delta_t}]\Big(m_t(\boldsymbol{x}-\boldsymbol{y}^j_0)  \\
& +m_{t-1}(\boldsymbol{y}^j_0-\boldsymbol{y}^i_0)+\sqrt{\delta_{t}}\boldsymbol{\epsilon}_t \Big).  \nonumber
\end{aligned}
\end{equation}
So the mean value can be simplified from Eq. (\ref{mu_ytj}) to the singe timestep expression as $\mu(\boldsymbol{y}^{js}_t, \boldsymbol{x})=c_{x t} \boldsymbol{x}+c_{y t} \boldsymbol{y}^{js}_t+c_{\epsilon t} (m_t(\boldsymbol{x}-\boldsymbol{y}^j_0)+m_{t-1}(\boldsymbol{y}^j_0-\boldsymbol{y}^i_0)+\sqrt{\delta_{t}}\boldsymbol{\epsilon}_t)$.
Compared with the mean value of prediction $p_\theta(\hat{\boldsymbol{y}}_{t-1}\mid \boldsymbol{y}_t, \boldsymbol{x})=\mathcal{N}(\hat{\boldsymbol{y}}^i_{t-1} ; \hat{\mu}_\theta(\boldsymbol{y}^i_t, \boldsymbol{x}, t), \delta_t \boldsymbol{I})$ as $\hat{\mu}_{\theta}(\boldsymbol{y}^i_t, \boldsymbol{x}, t)=c_{x t} \boldsymbol{x}+c_{y t} \boldsymbol{y}^i_t+c_{\epsilon t} \epsilon_\theta(\boldsymbol{y}^i_t, t)$, the gap between $\boldsymbol{y}^i_t$  and $\boldsymbol{y}^{js}_t$ is approximated to a single timestep and ignored. So the final training objective of inter-class loss can be formulated as:
\begin{equation}
\displaystyle
\scalebox{0.9}{$
\mathcal{L}_{inter} = \Vert m_t(\boldsymbol{x}-\boldsymbol{y}^j_0) + m_{t-1}(\boldsymbol{y}^j_0-\boldsymbol{y}^i_0) + \sqrt{\delta_t}\boldsymbol{\epsilon}_t -\epsilon_\theta(\boldsymbol{y}^i_t, t) \Vert_1. \nonumber
$}
\end{equation}

\section{C. Specified U-Net Structure Pruning}

\begin{figure}
\centerline{\includegraphics[width=0.99\linewidth]{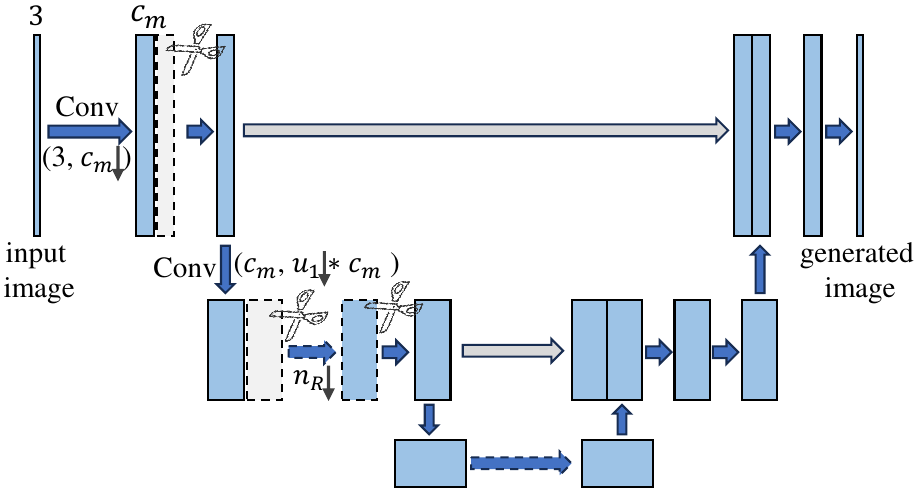}}
	\caption{The illustration of the pruning network architecture of U-Net. We take the contracting path (left side) as a schematic for the pruning process while the structure of the expansive path (right side) is adapted to match the adjusted contracting path.}
	\label{f-unet}
\end{figure}

\subsection{C.1 Pruning the structure of U-Net}

We give a schematic for the pruning process on the contracting path of the U-Net in Figure \ref{f-unet}, which shows different modules we care about to reduce the parameters of this network. It can be seen that It can be seen that three main factors influence the size of network parameters: the initial number of the model channels $C_m$, the multiplication factor for the increase in model channels $u=[u_1, u_2, ..., u_n]$ during the downsampling stage, and the number of ResNet blocks $n_R$ between two downsampling blocks.

We first decrease the initial model channel $C_m$ to $c_m=C_m/k$ where $k$ could adjust the degree of overall network pruning. For example, the reduction in model channels from 128 to 64 achieves a decrease in the parameters from 237.09M to 59.31M. Then we consider the upscale list $u=[u_1, u_2, ..., u_n]$ in downsampling modules and adjust it to lower multiplication factors, so the parameters in the corresponding modules can be effectively reduced. For example, by adjusting it from $[1,4,8]$ to a lower level such as $[1,4]$ or $[1,2,4]$ with $c_m=64$, the parameters can decrease from 59.31M to 13.00M or 16.22M. After adjusting $c_m$ and $u$, the channels of the network have been fully and effectively reduced, notice that there are still several ResNet blocks between two downsampling modules. So we cut this number $n_R$ from 2 to 1 with $c_m=64$ and $u=[1,4]$, the parameters can decrease from 13.00M to 9.39M. Combining the aforementioned pruning, we reduce the parameters from the initial BBDM model of 237.09M to 9.39M. Finally, we redesign the structure of the expansive path in the U-Net, which is adapted to match the adjusted contracting path, as shown in Figure \ref{f-unet}.

\subsection{C.2 Impact of different pruning modes}

We adjust the U-Net architecture and choose the original intra-class loss to optimize the transition. Table \ref{tab:cifar10block} shows the impact of different pruning modes for standard accuracy and adversarial robustness. There is a positive correlation between the adversarial robustness accuracy and the parameters. The robust performance of the model with 128 channels improves by 33.9\% and 47.3\% under PGD and AutoAttack, respectively, compared to the model with 32 channels. However, increasing $u=[1,4,8]$ when the initial model channels are 64 can further improve the performance under PGD and AutoAttack by 5\% and 5.4\%, respectively, with only a small increase in the parameters. Therefore, an initial channel setting of 64 offers more potential for robust performance and greater room for parameter reduction through pruning. In the detailed ablation of $c_m=64$, increasing the number of blocks $n_R$ has a minor improvement on the performance against AutoAttack, only by 0.1\%, but it achieves the highest performance under PGD attack. Changing $u$ from $[1,4]$ to $[1,2,4]$ incurs a low parameter cost, but significantly enhances performance against AutoAttack by 8\%, although there is a decline in standard accuracy and PGD attack performance. Overall, $u=[1,4]$ and $n_R=1$ maintain a relatively balanced standard and robust accuracy. We have reported the performance of this setting in the paper.

\begin{table}
    \centering
\resizebox{0.48\textwidth}{!}{
    \scriptsize
  \begin{tabular}{c|c|c|c|c|c|c}
    \toprule
    \multirow{2}*{$c_m$} & \multirow{2}*{$u$} &  \multirow{2}*{$n_R$} & \multirow{2}*{Params} & \multirow{2}*{S. Acc} & \multicolumn{2}{c}{Robust Acc} \\
     & &  &  &  & $\text{PGD}^{20}$ & AutoAttack \\
    \midrule
    \multirow{4}*{64} & $[1, 4]$ & 1 & 9.39M & 86.6 & 81.0 & 56.5 \\
    & $[1, 4]$ & 2 & 13M & 88.7 & \textbf{82.7} & 56.6 \\
    & $[1, 2, 4]$ & 1 & 11.82M & 85.8 & 80.4 & 64.5 \\
    & $[1, 4, 8]$ & 1 & 42.84M & 86.4 & 81.9 & \textbf{68.9} \\
    \midrule
    32 & $[1, 4]$ & 1 & 2.35M & 81.3 & 43.0 & 16.2 \\
    \midrule
    128 & $[1, 4]$ & 1 & 37.53M & 86.8 & 76.9 & 63.5 \\
    \bottomrule
  \end{tabular}
  }
    \caption{The ablation of the structure of U-Net used in IDC with PGD attack and AutoAttack $\ell_{\infty}$ ($\epsilon=8/255$) on CIFAR-10. '$c_m$' denotes the model channels, '$u$' is the upscale times and '$n_R$' is the number of the ResNet block. 'S. Acc' is an abbreviation for Standard accuracy.}
    \label{tab:cifar10block}
\end{table}

\begin{table}
    \centering
  \begin{tabular}{c|c|c|c}
    \toprule
    \multirow{2}*{$\alpha$} & \multirow{2}*{Standard Acc} & \multicolumn{2}{c}{Robust Acc} \\
      &  & $\text{PGD}^{20}$ & AutoAttack \\
    \midrule
    $0.15$ & 86.6 & 78.4 & 56.0 \\
    $[0.01, 0.2]$ & 87.0 & 84.8 & 57.6 \\
    $[0.01, 0.4]$  & 86.5 & 82.0 & \textbf{61.9} \\
    $0.2$ & 86.0 & \textbf{84.8} & 59.9 \\
    \bottomrule
  \end{tabular}
    \caption{Impact of loss weight $\alpha$ in our method on standard accuracy and robust accuracy against PGD attack and AutoAttack. The bracket $[\alpha_1,\alpha_2]$ represents the linear growth of alpha from $\alpha_1$ to $\alpha_2$ during training.}
    \label{tab:cifar10alpha}
\end{table}

\begin{figure*}
\centerline{\includegraphics[width=0.99\linewidth]{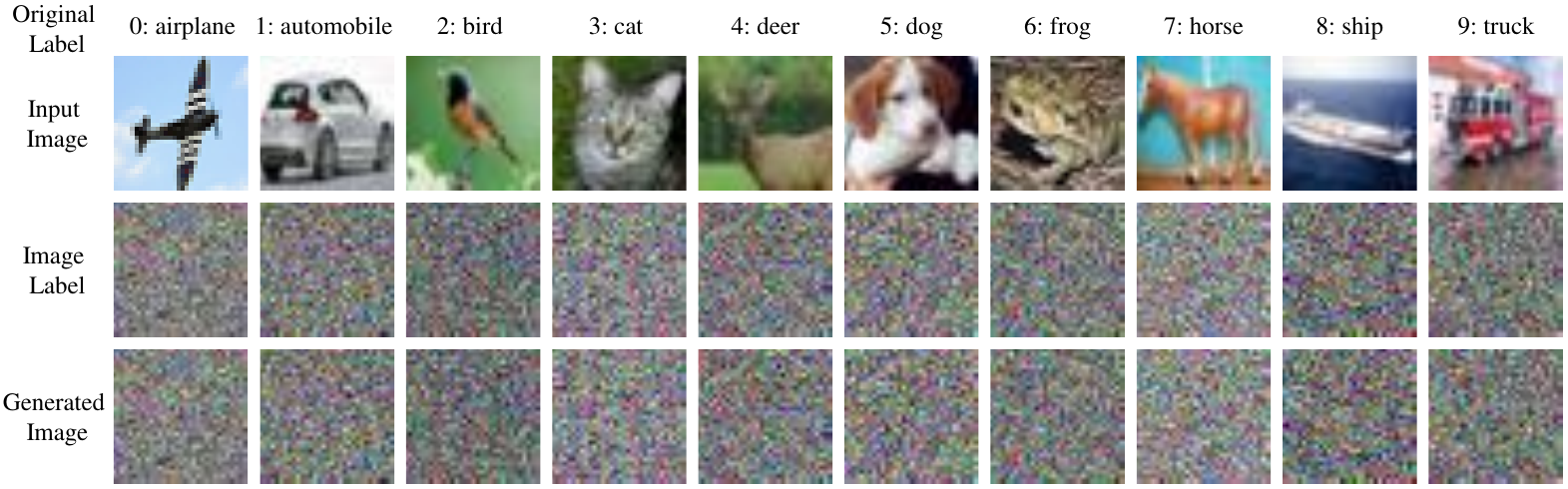}}
	\caption{Subjective presentation of input images under different labels, along with corresponding image labels and generated images. We randomly selected one image from 10 categories in the test dataset of CIFAR-10.}
	\label{f-subjective}
\end{figure*}

\subsection{C.3 Impact of loss weight $\bm{\alpha}$.}

Table \ref{tab:cifar10alpha} shows the ablation of the weight of the inter-class loss where we utilize the single value and linear values. By comparing different values of $\alpha$, it is evident that increasing $\alpha$ enhances the separability of the generated images from image labels of other classes, thereby improving the adversarial robustness. However, excessively large $\alpha$ may impose overly stringent constraints on distribution optimization, leading to decreased performance under PGD attacks. Although a linear increase benefits standard accuracy, it adversely affects the performance under AutoAttack. Consequently, we ultimately select $\alpha=0.2$ as the loss weight for our method.

\section{D. Subjective Results}

To better show the mechanism of the IDC work, we present in Figure \ref{f-subjective} the image labels corresponding to the input images and the generated images predicted by the IDC. The integer values in the first row are traditional labels, based on which a CNN-based classifier typically classifies. The natural images in the second row are randomly extracted from the CIFAR-10 dataset. The third row shows the image labels generated for the integer labels in the first row through our 'Orthogonal Image Labels Generation' module while we can perceive their differences subjectively. 
Our designed IDC can translate the natural images in the second row into the generated images in the fourth row through our designed 'Image-to-image Translation' framework, optimized by the 'classification loss' and the image labels in the third row. Subjectively, the noise color differences in the fourth row are essentially consistent with those in the third row, also demonstrating the effectiveness and rationality of our proposed framework.

\end{document}